\documentclass{article}

\PassOptionsToPackage{numbers, sort&compress}{natbib}
\usepackage[final]{neurips_2021_ml4ps}

\usepackage[utf8]{inputenc} 
\usepackage[T1]{fontenc}    
\usepackage{hyperref}       
\usepackage{url}            
\usepackage{booktabs}       
\usepackage{amsfonts}       
\usepackage{nicefrac}       
\usepackage{microtype}      
\usepackage{xcolor}         

\usepackage{graphicx}	
\usepackage{amsmath}	
\usepackage{amssymb}	
\usepackage{bm}
\usepackage[percent]{overpic}
\usepackage[normalem]{ulem}
\usepackage{tikz}
\usetikzlibrary{arrows,positioning}
\usepackage[font=small,labelfont=bf]{caption}
\usepackage{makecell}

\title{Real-time Detection of Anomalies in Multivariate Time Series of Astronomical Data}

\author{%
  Daniel Muthukrishna \\
  Massachusetts Institute of Technology\\
  Cambridge, MA, USA\\
  \texttt{danmuth@mit.edu} \\
   \And
   Kaisey S. Mandel \\
   University of Cambridge \\
   Cambridge, United Kingdom \\
   \texttt{kmandel@ast.cam.ac.uk} \\
   \And
   Michelle Lochner \\
   Department of Physics and Astronomy, University of the Western Cape\thanks{Bellville, Cape Town, 7535, South Africa} \\
   and South African Radio Astronomy Observatory (SARAO)\thanks{2 Fir Street, Observatory, Cape Town, 7925, South Africa} \\
   \texttt{mlochner@uwc.ac.za} \\
   \And
   Sara Webb \\
   Swinburne University of Technology \\
   Melbourne, VIC, Australia \\
   \texttt{swebb@swin.edu.au} \\
   \And
   Gautham Narayan \\
   University of Illinois at Urbana-Champaign \\
   Urbana, IL, USA \\
   \texttt{gsn@illinois.edu} \\
}

\begin{document}

\maketitle

\begin{abstract}
Astronomical transients are stellar objects that become temporarily brighter on various timescales and have led to some of the most significant discoveries in cosmology and astronomy. Some of these transients are the explosive deaths of stars known as supernovae while others are rare, exotic, or entirely new kinds of exciting stellar explosions. New astronomical sky surveys are observing unprecedented numbers of multi-wavelength transients, making standard approaches of visually identifying new and interesting transients infeasible. To meet this demand, we present two novel methods that aim to quickly and automatically detect anomalous transient light curves in real-time. Both methods are based on the simple idea that if the light curves from a known population of transients can be accurately modelled, any deviations from model predictions are likely anomalies. The first approach is a probabilistic neural network built using Temporal Convolutional Networks (TCNs) and the second is an interpretable Bayesian parametric model of a transient.  We show that the flexibility of neural networks, the attribute that makes them such a powerful tool for many regression tasks, is what makes them less suitable for anomaly detection when compared with our parametric model. 
\end{abstract}

\section{Introduction} 
Upcoming sky-surveys of the time-varying universe such as the Vera Rubin Observatory Legacy Survey of Space and Time (LSST) will observe over 10 million transient alerts each night: two orders of magnitude more than any survey to date \citep{Ivezic2009LSST:Products}. These new transient surveys will probe entirely new regimes in the time-varying universe, likely observing completely new classes of astronomical objects. For a long time, discovery in astronomy has been driven by serendipity \citep{Lang2009SerendipitousAstronomy} and by identifying anomalies in datasets, which has primarily been achieved by visual examination and follow-up observations. However, with the large influx of data, it is infeasible to visually examine or follow up any significant fraction of transient candidates. To this end, identifying anomalous objects and prioritising which of the millions of alerts are most suitable for followup observations is a challenge that needs to be automated. In this paper, we develop a novel framework for identifying anomalous transients in real-time.

Most previous efforts to automate the identification of transients require the full phase coverage of the transient. While retrospective classification after the full light curve of an event has been observed is useful, it also limits the scientific questions that can be answered about these events, many of which exhibit interesting physics at early times. There has been very little focus at identifying transients in real-time. Obtaining detailed followup observations shortly after a transient's explosion provides insights into the object's physical mechanism. While the mechanism of some transients are reasonably well-understood, the central engine of various exotic classes are poorly understood \citep[e.g.][]{CoppejansMargutti2020FBOTs}.

\section{Method}
To evaluate our method, we used the dataset described in \citet{Muthukrishna19RAPID} that augmented real transients \citep{KesslerPlasticcModels} using the \texttt{SNANA} software \citep{Kessler2010SNANA:Analysis} to match the observing properties of the Zwicky Transient Facility (ZTF) \citep{Bellm2019ZTF}. We simulated 10,000 transients from three common transient classes (SNIa, SNII, SNIbc) and eight rare transient classes (Kilonova, SLSN-I, TDE, CART, PISN, ILOT, AGN, uLens-BSR). Each simulated transient consists of a time-series of flux and flux uncertainty measurements in the $g$ (green wavelength range) and $r$ (red wavelength range) passband filters, known as a \textit{light curve}.

Our methods for anomaly detection involve first developing an autoregressive sequence model of a transient class, and then using the model's ability to predict future fluxes as an anomaly score (or conversely, a goodness of fit score). We develop two methods for regressing over a transient. The first is a probabilistic deep neural network (DNN) approach using Temporal Convolutional Networks (TCNs), and the second is a Bayesian parametric approach using the flexible Bazin function \cite{Bazin_function}. 

Each model aims to do real-time detection, and is hence causal, using only past values to predict future values. Specifically, our model is a function that predicts future fluxes in a time-series as well as the uncertainty of that prediction; it then compares the prediction with the observed data to obtain an anomaly score. 

In the following two subsections (\S \ref{sec:Model_NN}, \S \ref{sec:Model_Bazin}), we describe our two approaches of developing a function that maps the observed fluxes for a transient $s$ at passband $p$ up to time $T$ onto flux predictions $y_{sp(T+3)}$ and predictive uncertainties $\sigma_{y,{sp(T+3)}}$ three days after a given set of observations:
\begin{minipage}{\textwidth}
    \begin{minipage}[b]{0.65\textwidth}
        \begin{tikzpicture}[node distance = 6mm and 12mm, align=center]
        \node (adc) [draw,minimum size=14mm] {Model \\(DNN or Bazin)};
        
        \coordinate[above left = 3mm and 12mm of adc.west]   (a1);
        \coordinate[below = of a1]              (a2);
        \coordinate[below = of a2]              (a3);
        
        \coordinate[above right= 3mm and 12mm of adc.east]  (b1);
        \coordinate[below=of b1]  (b2);
        
        \draw[-latex']  (a1) node[left] {$\bm{D}_{sp(t \le T)}$} -- (a1-| adc.west);
        \draw[-latex']  (a2) node[left] {$\bm{\sigma}_{D,{sp(t \le T)}}$} -- (a2-| adc.west);
        
        \draw[-latex'] (adc.east |- b1) -- (b1) node[right] {$y_{sp(T+3)}$};
        \draw[-latex'] (adc.east |- b2) -- (b2) node[right] {$\sigma_{y,{sp(T+3)}}$};
        
        \end{tikzpicture}
    \end{minipage}
    \begin{minipage}[b]{0.34\textwidth}
        \begin{align}
            \bm{X}_{spt} &= \left[D_{spt}, \sigma_{D,{spt}} \right] 
            \label{eq:X_input}
        \end{align}
        \begin{align}
            \bm{Y}_{spt} &= \left[y_{spt}, \sigma_{y,{spt}} \right]
            \label{eq:Y_predictions}
        \end{align}
    \end{minipage}
\end{minipage}

The DNN approach builds a neural network that effectively performs regression over past data in order to predict the flux 3 days in the future. On the other hand, the Bazin approach performs regression over time to predict the flux at any time. We then feed in partial light curves into the Bazin model and infer a prediction 3 days after given data to obtain anomaly scores comparable with the DNN. We choose to predict the flux 3 days in the future to approximately match the 3-day cadence between ZTF observations.

\subsection{Probabilistic Neural Network}
\label{sec:Model_NN}

\begin{figure*} 
    \centering
    \includegraphics[width=1\linewidth]{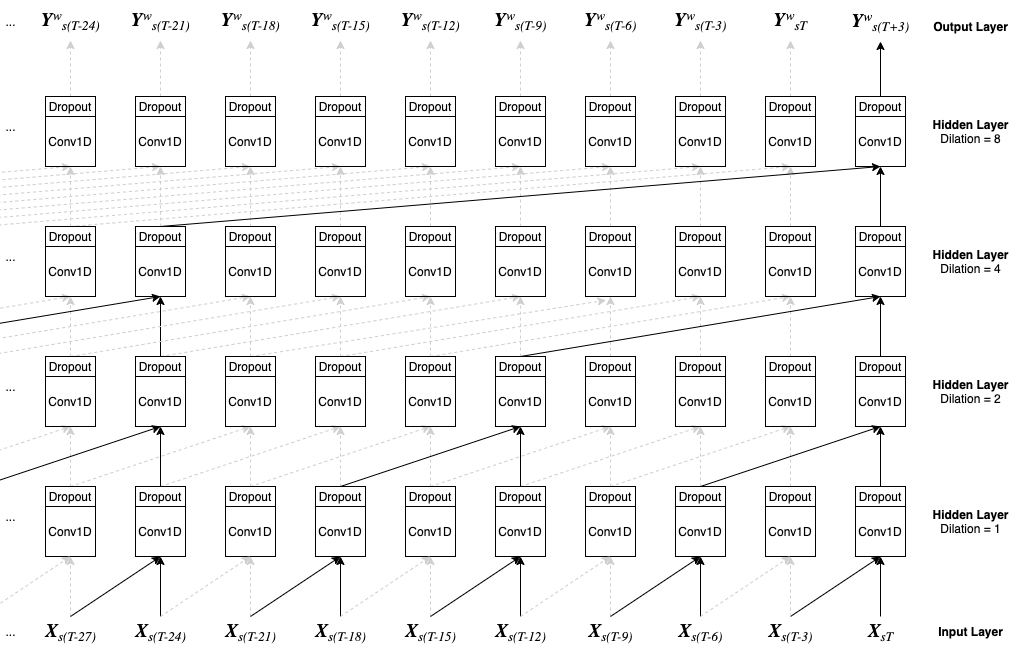}
    \caption{Temporal Convolutional Neural Network architecture used in this work. Each column in the diagram is a subsequent time-step from left to right. The bottom row is the input light curve (from equation \ref{eq:X_input}) where each input is a vector of the observed flux and flux uncertainty in all passbands for a transient $s$ at a time $t$. The input fluxes and uncertainties of two adjacent time-steps are passed into a residual block consisting of a 1D convolutional neural network layer (Conv1D) with dropout. While not shown in the figure, the residual block also contains a second Conv1D layer with dropout. The outputs of these are then convolved with the outputs from some previous time-steps in the above hidden layers as shown in the diagram, until the final Output Layer is the predicted light curve at the following time-step (from equation \ref{eq:Y_predictions}). The solid arrows show how the prediction $\bm{Y}^w_{s(T+3)}$ is made, and the gray dashed arrows show the neural network layers that lead to all other predictions. The network is causal, whereby new predictions only use information from previous time-steps in the light curve. We set the dropout rate to $20\%$ for all layers in the network. We build this model using the \texttt{Keras} and \texttt{TensorFlow Probability} libraries after adapting the TCN model from \citet{Bai2018} and their code in \url{https://github.com/philipperemy/keras-tcn}.}
    \label{fig:tcn_architecture}
\end{figure*}

The DNN is an autoregressive mapping function that aims to map an input multi-passband light curve matrix, $\bm{X}_{s(t \le T)}$, for transient $s$ up to a time $T$, onto an output multi-passband flux vector at the next time-step $T+3$,
\begin{equation}
    \bm{Y}^w_{s(T+3)} = \bm{f}_T(\bm{X}_{s(t\le T)}; \bm{w})
    \label{eq:learned_function}
\end{equation}
where $\bm{w}$ are the parameters (i.e. weights and biases) of the network. We define $\bm{X}_{s(t \le T)}$ as the matrix $\bm{X}_s$ but up to a time $T$ in each of its $N_p$ passbands. The model output prediction $\bm{Y}^w_{s(T+3)}$ is a $1 \times 2N_p$ vector consisting of the predicted mean flux $\tilde{y}_{sp(T+3)} (\bm{w})$ and intrinsic uncertainty $\tilde{\sigma}_{\mathrm{int},sp(T+3)}(\bm{w})$ in the $g$ and $r$ passbands at the next time-step for a particular set of network weights. The model $\bm{f}_T(\bm{X}_{s(t \le T)}; \bm{w})$ in equation \ref{eq:learned_function} is represented by the complex DNN architecture illustrated in Fig.~\ref{fig:tcn_architecture}.

To characterise the intrinsic uncertainty of our network's ability to represent a light curve, we build a probabilistic neural network. Our DNN parameterizes a Normal distribution and outputs a predictive mean $\tilde{y}_{spt}(\bm{w})$ and standard deviation $\tilde{\sigma}_{\mathrm{int},spt}(\bm{w})$ for a particular set of network weights $w$. The predictions $\tilde{y}_{spt}(\bm{w})$ and $\tilde{\sigma}_{\mathrm{int},spt}(\bm{w})$ are components of the vector $\bm{Y}^w_{spt}$ from equation \ref{eq:learned_function}. We include the learned uncertainty $\tilde{\sigma}_{\mathrm{int},spt} (\bm{w})$ because we know that our DNN model is not a perfect representation of a light curve, and even if we had no measurement error and had an infinite training set, there would still be some discrepancy between our DNN predictions and the observed light curves. 

A Bayesian neural network enables us to also quantify the uncertainty in our model's predictions of the outputs, $\tilde{y}_{spt}(\bm{w})$ and $\tilde{\sigma}_{\mathrm{int},spt}(\bm{w})$. We approximate a Bayesian network using an approach called Monte Carlo (MC) dropout sampling (see \citet{Gal2015}). We use MC dropout with our probabilistic neural network to estimate the predictive uncertainty. We do this by collecting the results of stochastic forward passes through the network as approximate posterior draws of $\bm{Y}^w_{spt}$, and use the mean and standard deviation of these draws as our marginal predictive mean $y_{spt}$ and predictive uncertainty $\sigma_{y,{spt}}$. We define the model loss function as a log-likelihood comparing the flux predictions and observations including model and data uncertainties as described in \ref{sec:NN_loss_function}.

\subsection{Parametric Bayesian Bazin function}
\label{sec:Model_Bazin}
We have also built a Bayesian model of each transient light curve based on the widely used phenomenological Bazin function from \citet{Bazin_function},
\begin{equation}
    f_{spt}(\bm{\theta}) = A \frac{e^{-(t-t_0)/\tau_{\mathrm{fall}}}}{1 + e^{-(t-t_0)/\tau_{\mathrm{rise}}}} + B
\label{eq:bazin_function_no_errors}
\end{equation}
where $\bm{\theta} = [A, B, t_0, \tau_{\mathrm{fall}}, \tau_{\mathrm{rise}}, \sigma_{\mathrm{int}}]$ are the free parameters of the model. We then model the latent flux of a transient $s$ in passband $p$ with the Bazin function plus an additional intrinsic error $\epsilon_{\mathrm{int}}(t)$ to account for discrepancies between the model and the observed light curve, 
\begin{equation}
    F_{spt}(\bm{\theta}) = f_{spt}(\bm{\theta}) + A\epsilon_{\mathrm{int}}(t),
    \label{eq:bazin_function}
\end{equation}
where $\epsilon_{\mathrm{int}}(t) \sim \mathcal{N}(0,\sigma_{\mathrm{int}}^2)$ is a zero-mean Gaussian random variable with intrinsic variance $\sigma_{\mathrm{int}}^2$. Next, we derive a generative model of the observed flux, $D_{spt} = F_{spt}(\bm{\theta}) + \epsilon_{D,{spt}}$, where we assume that the measurement error $\epsilon_{D_{spt}} \sim \mathcal{N}(0,\sigma_{D,{spt}}^2)$ is a zero-mean Gaussian random variable with variance $\sigma_{D,{spt}}^2$.

To model a partial light curve from -70 days before the first detection (\textit{trigger}) up to time $T$, we write the likelihood function as follows,
\begin{equation}
\begin{split}
    &\mathcal{P}(\bm{D}_{sp(t \le T)}|\bm{t}, \bm{\theta}) =  \prod^{T}_{t=-70} \mathcal{N} \left(  D_{spt}  \mid f_{spt}(\bm{\theta}), A^2\sigma_{\mathrm{int}}^2 +  \sigma_{D,{spt}}^2\right)
\end{split}
\label{eq:bazin_likelihood}
\end{equation}

We define a Bayesian model to fit each transient light curve in a particular passband after choosing a Gaussian prior $\mathcal{P}(\bm{\theta})$ based on the population of transients, $\mathcal{P}(\bm{\theta}|\bm{D}_{sp},\bm{t}) \propto \mathcal{P}(\bm{D}_{sp}|\bm{t}, \bm{\theta}) \mathcal{P}(\bm{\theta})$. In practice, for anomaly detection, we need to make predictions of $D$ at new times $T$. This requires that we evaluate the predictive distribution defined by
\begin{equation}
    \mathcal{P}(F_{sp(T+3)} | \bm{D}_{sp(t \le T)}, \bm{t}) = \int \mathcal{P}(F_{sp(T+3)}|\bm{\theta}, \bm{t}) \mathcal{P}(\bm{\theta}|\bm{D}_{sp(t \le T)}, \bm{t}) d\bm{\theta}.
\label{eq:Bazin_predictive_distribution}
\end{equation}

The integral on the RHS cannot be computed analytically, and so we approximate it by sampling. We draw sample parameters of the posterior (second term in the integrand) and compute the flux predictions for each set of parameters (first term in the integrand) with equation \ref{eq:bazin_function}. The LHS of equation \ref{eq:Bazin_predictive_distribution} is the sampled probability density function and we estimate its marginal predictive mean and uncertainty as the sample mean and standard deviation of the fluxes computed from the posterior draws.

\section{Results and Discussion}
We trained three autoregressive DNN models, one for each common transient class: SNIa, SNII, SNIbc. And similarly, for the Bazin function, we defined a prior distribution for each of these common classes based on the population of transients in each training set. Each training set consisted of $\sim$8000 light curves and we validated the performance of the models on $\sim$2000 light curves from each transient class. We then applied our models to the transients from all other classes to identify how well the rare transient classes were identified as anomalous.

\begin{figure*}[ht]
    \centering
    \includegraphics[width=0.49\linewidth]{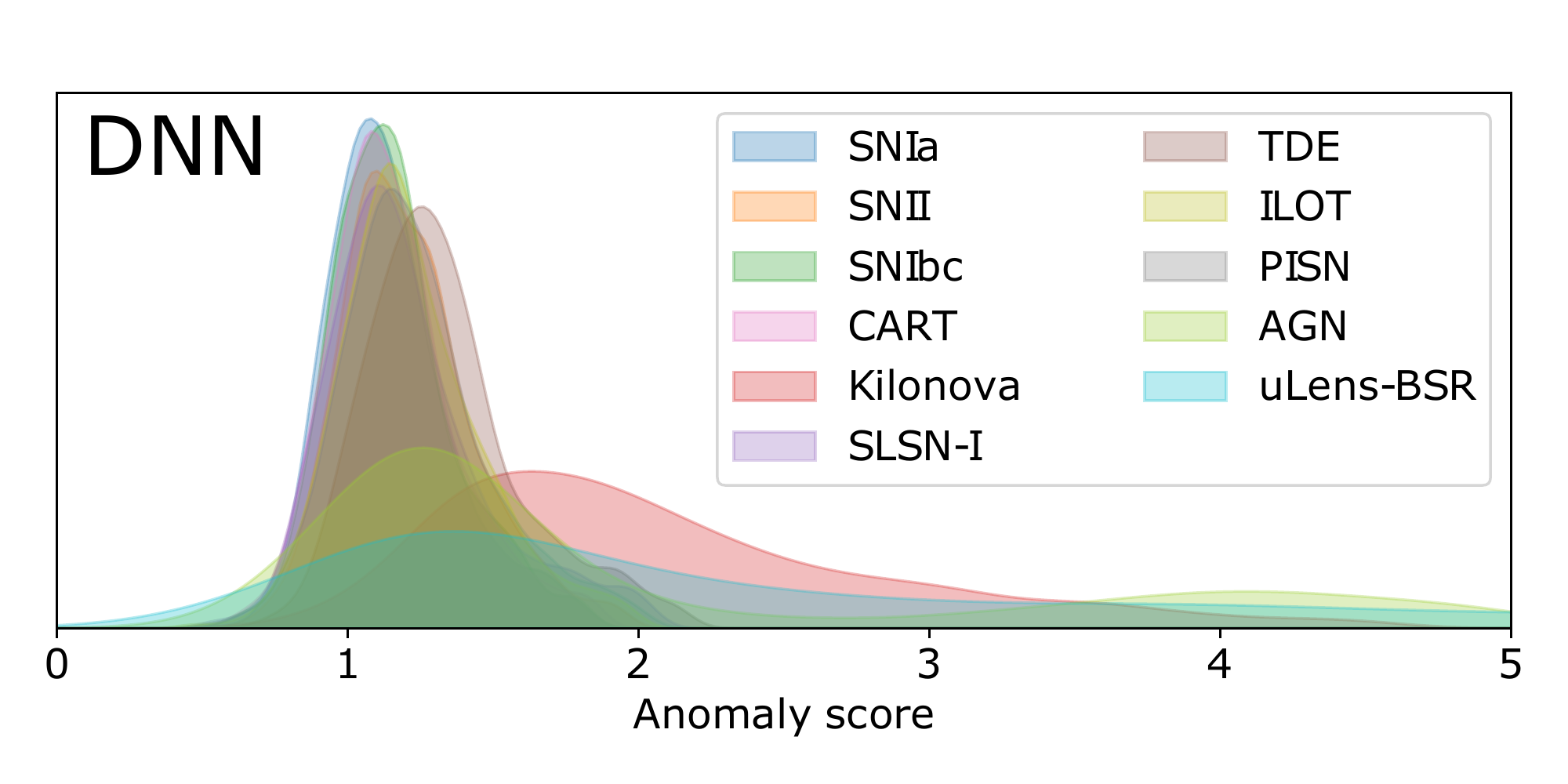}
    \includegraphics[width=0.49\linewidth]{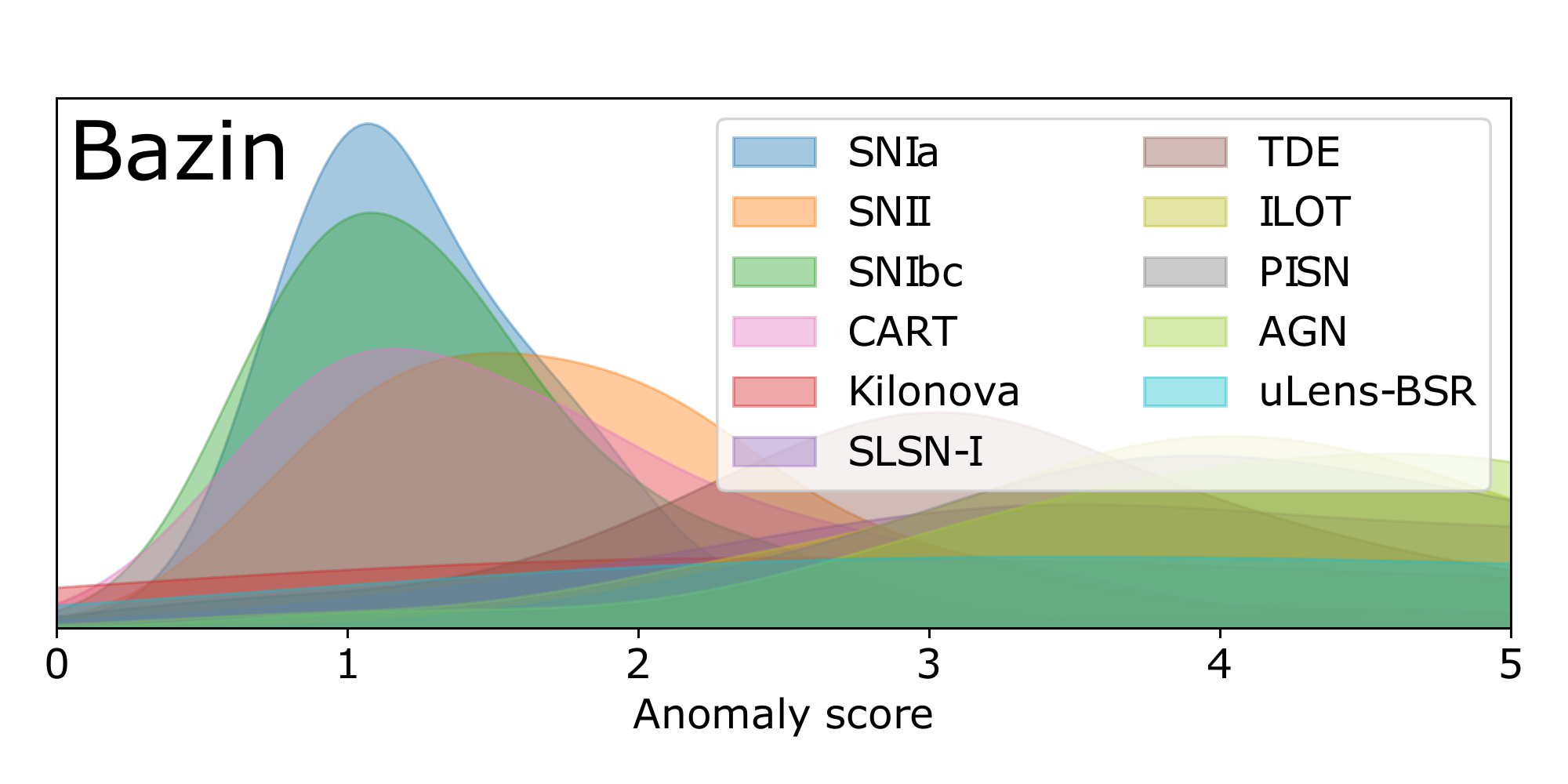}
    \caption{Anomaly score distribution recorded over the full light curve for the models that were trained on SNIa light curves and tested on the transient populations of ten different classes. Classes that are dissimilar to SNIa have higher anomaly scores, while similar classes have lower anomaly score distributions. The Bazin plot (right) shows a larger separation of the distributions of the anomalous classes from the common transient classes (SNIa, SNII, SNIbc) than the DNN (left), indicating that it is better at identifying anomalous classes. We refer the reader to \citet{Muthukrishna2021} for a more comprehensive comparison and analysis of the two methods.}
    \label{fig:Anomaly_score_distribution}
\end{figure*}

We quantify anomaly scores using a $\chi^2$ metric to compute the discrepancy between the observed flux at time $t$ and the predictions of a model based on previous data. This $\chi^2$ is weighted by the total variance including the predictive uncertainty and measurement error. A large $\chi^2$ occurs when the model does not predict observations well and can thus be indicative of an anomaly. As illustrated in Figure \ref{fig:Anomaly_score_distribution}, the Bazin model is significantly better at identifying anomalous classes than the DNN models. In Appendix \ref{sec:Appendix_Comparison_of_DNN_and_Bazin}, we highlight that the poor performance of the DNN compared to the Bazin model is because it is too flexible at predicting light curves; and after being trained on one class, it is still able to accurately predict fluxes in a different class of transients. The DNN model is actually better at predicting the future fluxes of transients within a trained class, but is also able to predict the future fluxes of transients from different classes well. While this flexibility allows for good flux predictions, it is not good for anomaly detection. 

\begin{figure}[ht]
    \centering
    \includegraphics[width=0.49\linewidth]{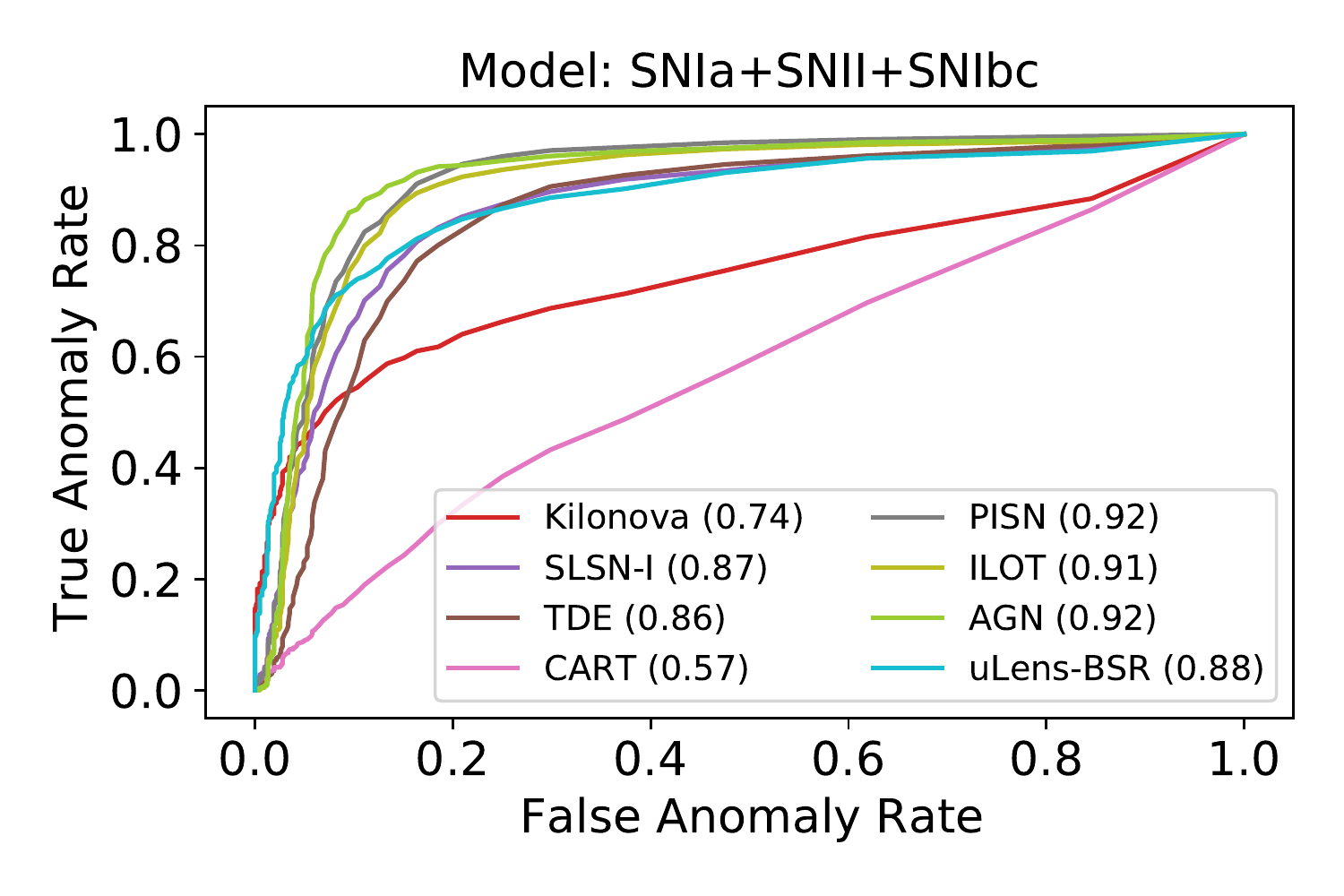}
    \includegraphics[width=0.49\linewidth]{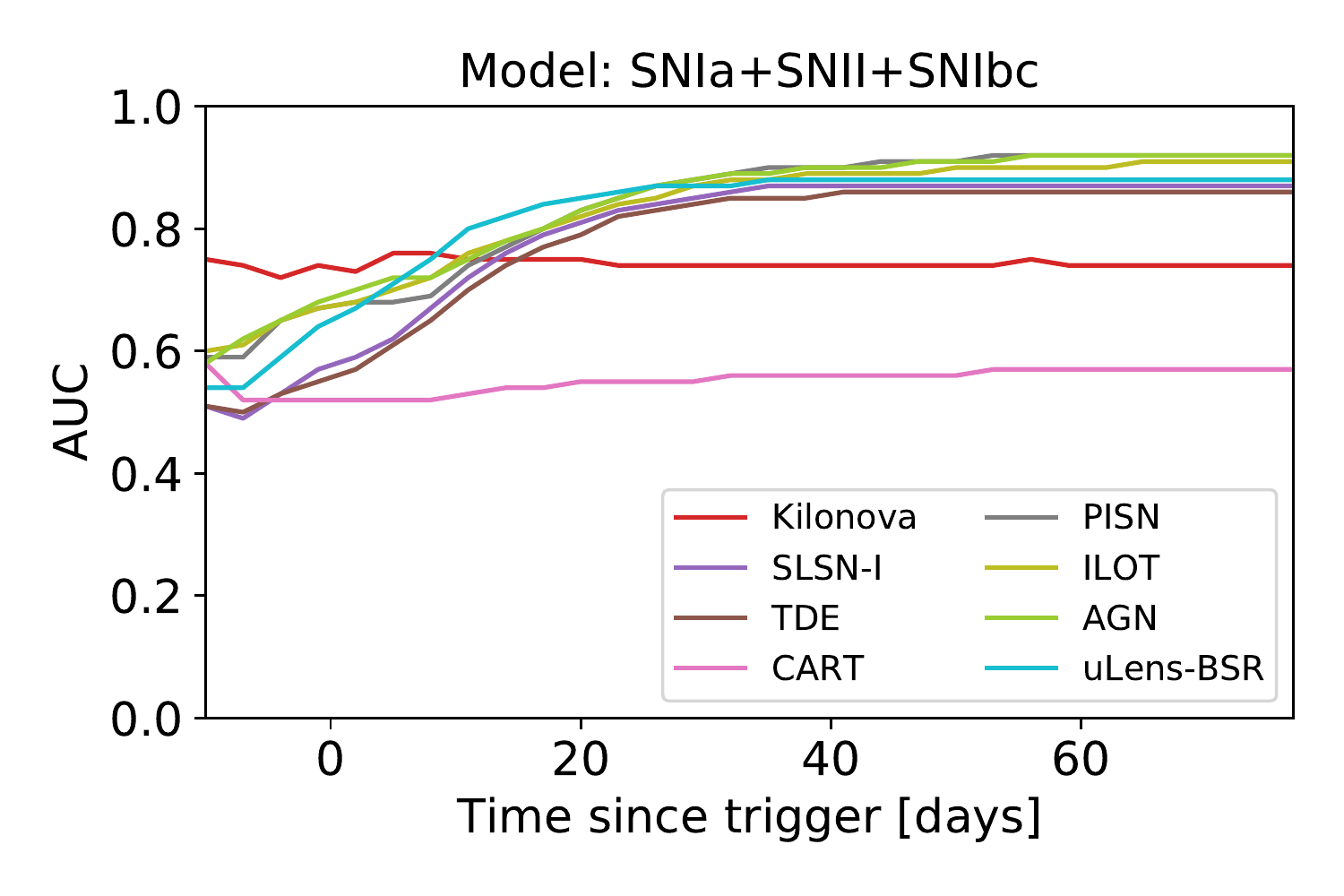}
    \caption{\textit{Left panel:} The Receiver Operating Characteristic (ROC) curve plotting the True Anomaly Rate against the False Anomaly Rate for different threshold anomaly scores. We aggregated the results of the Bazin models trained on each of the common transients (SNIa, SNII and SNIbc) and show their combined performance on eight anomalous classes denoted in the legend. The area under the curves (AUCs) are shown in brackets in the legend. We use the anomaly scores over the full light curves to make these ROC curves. \textit{Right panel:} We aggregate the ROC curves at all times by plotting the AUC scores as the light curves evolve, illustrating that our ability to identify anomalies improves with time. The model performs very well at distinguishing all classes except for CARTs which are known to be difficult to distinguish from common SNe based only on the light curves \citep{Muthukrishna19RAPID}.
    }
    \label{fig:ROC_curve_SNIaSNIISNIbc}
\end{figure}

Our methods allow us to identify anomalies as a function of time, and in Figure \ref{fig:ROC_curve_SNIaSNIISNIbc}, we show that the Bazin model is able to have high True Anomaly Rates and low False Anomaly Rates that improve over the lifetime of a transient, reaching AUC scores well above 0.8 for most rare classes. The DNN model, on the other hand, achieves an average AUC score of only 0.66 across the rare classes. In future work, we hope to apply our method on real-time ZTF observations to gauge our success at identifying real anomalous transients. In practice, applying an anomaly detection framework in conjunction with a transient classifier will provide more valuable information on whether a newly discovered transient is interesting enough for further follow-up observations. An issue with this work, is that there has been no distinction between anomalies and \textit{interesting} anomalies. We expect that most bogus transient alerts will already be removed by \textit{real-bogus} cuts \citep[e.g.][]{Duev2019}, however, our approach may still flag unusual transient phenomena that don't align with our trained transient classes but are uninteresting to most astronomers. To deal with this, future work should consider Active Learning frameworks that use methods such as \textit{Human-in-the-loop learning} that specifically target what users define as interesting phenomena (recent work by \citet{Ishida2019_Timeseries,Lochner2020Astronomaly} have begun working on Active Learning for anomaly detection).

Overall, this paper presents a novel and effective method of identifying anomalous transients in real-time that is fast enough to be easily scaleable to surveys as large as LSST. Anomaly detection coupled with other classification approaches enables astronomers to prioritise follow-up candidates. This work and other recent approaches to transient anomaly detection \citep[e.g.][]{Villar2021_Anomalydetection,Pruzhinskaya2019,Malanchev2021,Webb2020,Soraism2020Novelties,Muthukrishna2021} are going to be critical for discovery in the new era of large-scale astronomical surveys.

\medskip

\begin{ack}
ML acknowledges support from South African Radio Astronomy Observatory and the National Research Foundation (NRF) towards this research. Opinions expressed and conclusions arrived at, are those of the authors and are not necessarily to be attributed to the NRF. KSM acknowledges funding from the European Research Council under the European Union's Horizon 2020 research and innovation programme (ERC Grant Agreement No. 101002652). This project has been made possible through the ASTROSTAT-II collaboration, enabled by the Horizon 2020, EU Grant Agreement No. 873089.

\end{ack}

\small
\bibliography{references}

\appendix

\section{Appendix}

\subsection{DNN Model loss function}
\label{sec:NN_loss_function}
Before defining the loss function, we first develop a generative model of the latent flux of a transient $s$ in passband $p$ 3 days in the future at time $T+3$. We aim to model the underlying latent flux with the neural network as follows,
\begin{equation}
   F_{sp(T+3)}(\bm{w}) = \tilde{y}_{sp(T+3)}(\bm{w}) + \epsilon_{\mathrm{int},sp(T+3)}(\bm{w}),
\label{eq:DNN_latent_flux}
\end{equation}
where the error $\epsilon_{\mathrm{int},sp(T+3)}(\bm{w}) \sim \mathcal{N}\left(0, \tilde{\sigma}^2_{\mathrm{int},sp(T+3)}(\bm{w})\right)$ is a zero-mean Gaussian random variable with variance $\tilde{\sigma}^2_{\mathrm{int},sp(T+3)}(\bm{w})$.
Thus, we write the predictive distribution of the latent flux as follows,
\begin{equation}
    \begin{split}
      \mathcal{P}(\bm{F}_{s(T+3)}|\bm{X}_{s(t \le T)}, \bm{w}) = \prod^{N_p}_{p=1} \mathcal{N} \left( F_{sp(T+3)}(\bm{w}) \mid \tilde{y}_{sp(T+3)}(\bm{w}), \tilde{\sigma}^2_{\mathrm{int},sp(T+3)}(\bm{w}) \right).
    \end{split}
\label{eq:posterior_predictive}
\end{equation}
Next, a generative model of the observed flux is derived by adding a measurement error to the latent flux as follows,
\begin{equation}
    D_{sp(T+3)} = F_{sp(T+3)}(\bm{w}) + \epsilon_{D, sp(T+3)},
\label{eq:generative_model_data_DNN}
\end{equation}
where we assume that the measurement error $\epsilon_{D, sp(T+3)} \sim \mathcal{N}(0,\sigma^2_{D,{sp(T+3)}})$ is a zero-mean Gaussian random variable with variance $\sigma^2_{D,{sp(T+3)}}$.

Typically, researchers will not use the uncertainty in the data within the loss function (e.g. \citet{JamalBloom2020,Villar2021_Anomalydetection}, however, work by \citet{Naul2018AStars} included data uncertainty without model uncertainty). In this work, we construct our loss function to include both predictive uncertainties $\tilde{\sigma}_{\mathrm{int},sp(T+3)}(\bm{w})$ and flux uncertainties $\sigma_{D,{sp(T+3)}}$.
Given equations \ref{eq:DNN_latent_flux} and \ref{eq:generative_model_data_DNN}, we write the likelihood function of the probabilistic DNN as follows,
\begin{equation}
    \begin{split}
      \mathcal{P}(\bm{D}_{s(T+3)}|\bm{X}_{s(t \le T)}, \bm{w}) &= \prod^{N_p}_{p=1} \mathcal{N} \left( D_{sp(T+3)} \mid \tilde{y}_{sp(T+3)}(\bm{w}), \tilde{\sigma}^2_{\mathrm{int},sp(T+3)}(\bm{w}) + \sigma_{D,{sp(T+3)}}^2  \right) \\
            &= \prod^{N_p}_{p=1} \left( 2\pi(\tilde{\sigma}^2_{\mathrm{int},sp(T+3)}(\bm{w}) + \sigma_{D,{sp(T+3)}}^2) \right)^{-0.5} \\
            &\quad \times \exp{\left( -0.5 \frac{(\tilde{y}_{sp(T+3)}(\bm{w})-D_{sp(T+3)})^2}{\tilde{\sigma}^2_{\mathrm{int},sp(T+3)}(\bm{w}) + \sigma_{D,{sp(T+3)}}^2} \right) }.
    \end{split}
\label{eq:DNN_likelihood}
\end{equation}

Following \citet{Gal2015}, we define the prior over the weights as a zero-mean Normal distribution,
\begin{equation}
    \mathcal{P}(\bm{w}) = \mathcal{N}(\bm{w} \mid 0, \bm{I}/l^2),
\label{eq:DNN_prior}
\end{equation}
where $\bm{I}$ is the identity matrix and $l$ is the prior length-scale that regularises how large the weights can be. The posterior over the weights is given by the product of the prior distribution and the likelihood function over all $N_s$ transients at all $N_t$ time-steps,
\begin{equation}
     \mathcal{P}(\bm{w} |\bm{X}) \propto \mathcal{P}(\bm{w}) \prod\limits_{s=1}^{N_s}\prod\limits_{T=-70}^{80} \mathcal{P}(\bm{D}_{s(T+3)}|\bm{X}_{s(t \le T)}, \bm{w}),
\label{eq:DNN_posterior}
\end{equation}
where we ignore the Bayesian evidence as a scaling constant that is unnecessary for this work. We would ideally like to sample the negative log posterior while training our DNN, and so we derive the log prior from equation \ref{eq:DNN_prior} as $\log{\mathcal{P}(\bm{w})} = \mathrm{constant} - l^2 ||\bm{w}||_2^2 / 2$. We can ignore the additive constant not necessary for our optimisation and follow \citet{Gal2015} to implement the log prior by including an $L_2$ regularisation term $\lambda ||\bm{w}||_2^2$ weighted by some weight decay that averages over the number of transients $N_s$ and time-steps $N_t$,
\begin{equation}
    \lambda = \frac{l^2 (1-d)}{2 N_s N_t},
\end{equation}
where $d$ is the dropout rate (set to $0.2$ in this work), and we set $l=0.2$ consistent with work by \citet{Gal2015}\footnote{See Section 4.2 of \citet{Gal2015Appendix} for a detailed explanation of this prior and \url{https://github.com/yaringal/DropoutUncertaintyExps/blob/master/net/net.py} for an example implementation of this $L_2$ regularisation by Yarin Gal.}. Here, we have included the $(1-d)$ term to account for the dropout regularisation used in our work. With this term, the $L_2$ regularisation term $\lambda ||\bm{w}||_2^2$ matches the log prior, $\log{\mathcal{P}(\bm{w})}$, averaged over the number of transients and time-steps. However, we point out that our $\lambda$ slightly differs from equation 18 of \citet{Gal2015Appendix} because we use the negative log-likelihood instead of a squared loss as the cost function of our DNN. Furthermore, we also add the caveat that because of this difference of loss functions and our inclusion of a probabilistic neural network that outputs a predictive mean and standard deviation instead of a point estimate, it is not clear that the demonstration of MC dropout as an approximation to Bayesian neural networks in \citet{Gal2015} necessarily holds true in our work. Future machine learning research should check the validity of MC dropout as a Bayesian approximation in a broader range of neural network architectures.



Since we use dropout regularisation, we define a dropout objective function over all time-steps and over all transients that we aim to minimise while training the neural network model as follows,
\begin{equation}
     \mathrm{obj}(\bm{w}) = \sum\limits_{s=1}^{N_s}\sum\limits_{T=-70}^{80} 
     \left[ -\log \mathcal{P}(\bm{D}_{s(T+3)}|\bm{X}_{s(t \le T)}, \bm{w}) + \lambda ||\bm{w}||_2^2 \right]
    \label{eq:objective}
\end{equation}
where we sum the log-likelihood and $L_2$ regularisation term over all $N_t$ time-steps (between -70 and 80 days) and $N_s$ transients in the training set. To train the DNN and determine optimal values of its parameters $\bm{\hat{w}}$, we minimise the dropout objective function with the sophisticated and commonly used \texttt{Adam} gradient descent optimiser \citep{Kingma2014}. 

To make predictions, we evaluate the predictive distribution of the latent flux defined as follows,
\begin{equation}
    \mathcal{P}(\bm{F}_{s(T+3)} | \bm{X}_{s(t \le T)}) = \int \mathcal{P}(\bm{F}_{s(T+3)} | \bm{X}_{s(t \le T)}, \bm{w}) \mathcal{P}(\bm{w} |\bm{X}) d\bm{w},
    \label{eq:DNN_predicitve_distribution}
\end{equation}
where we are marginalising over the weights of the network by integrating the product of the predictive distribution of the latent flux given the network weights (first term in the integrand and defined in equation \ref{eq:posterior_predictive}) and the posterior distribution over the network weights (second term in the integrand and defined in equation \ref{eq:DNN_posterior}). The integral is intractable, and so we approximate it by using Monte Carlo dropout at inference time to sample the posterior distribution, as described in \cite{Gal2015}. We draw 100 samples from the posterior $\mathcal{P}(\bm{w} |\bm{X})$ by running 100 forward passes of the neural network for a given input. Each run of the neural network outputs both a mean $\tilde{y}_{sp(T+3)}(\bm{w}_\mathrm{draw})$ and standard deviation $\tilde{\sigma}_{\mathrm{int},sp(T+3)}(\bm{w}_\mathrm{draw})$ because of our probabilistic neural network architecture. To include the variance of each draw in the marginal predictive uncertainty $\sigma_{y,{sp(T+3)}}$, we compute $F_{sp(T+3)}(\bm{w}_\mathrm{draw}) \sim \mathcal{N} \left( \tilde{y}_{sp(T+3)}(\bm{w}_\mathrm{draw}), \tilde{\sigma}^2_{\mathrm{int},sp(T+3)}(\bm{w}_\mathrm{draw}) \right)$. We estimate the marginal predictive mean and uncertainty as the sample mean and standard deviation of the 100 values of $F_{sp(T+3)}(\bm{w}_\mathrm{draw})$ taken from the 100 forward passes of the neural network, respectively:
\begin{equation}
    y_{sp(T+3)} = \frac{1}{100} \sum_{\mathrm{draw=1}}^{100} F_{sp(T+3)}(\bm{w}_\mathrm{draw}),
\end{equation}
\begin{equation}
    \sigma_{y,{sp(T+3)}}  = \sqrt{\frac{1}{100} \sum_{\mathrm{draw=1}}^{100} \left(F_{sp(T+3)}(\bm{w}_\mathrm{draw}) - y_{sp(T+3)} \right)^2}.
\end{equation}
These outputs are used to compute the anomaly scores.

\subsection{Comparison of DNN and Bazin predictive power}
\label{sec:Appendix_Comparison_of_DNN_and_Bazin}
    We performed the following analysis to evaluate why the DNN model was less effective at identifying anomalies than the Bazin model. To compare the models on an even scale, we defined the Measurement-Uncertainty-Scaled Prediction Error as follows,
        \begin{equation}
                \mathrm{MUSPE}_{spt} = \frac{(y_{spt} - D_{spt})} {\sigma_{D,{spt}}}.
        \label{eq:physical_flux_error}
        \end{equation}

\begin{figure*}[ht]
        \centering
        \textbf{Bazin}\par\medskip
        \vspace{-0.7em}
        \includegraphics[width=1.0\linewidth]{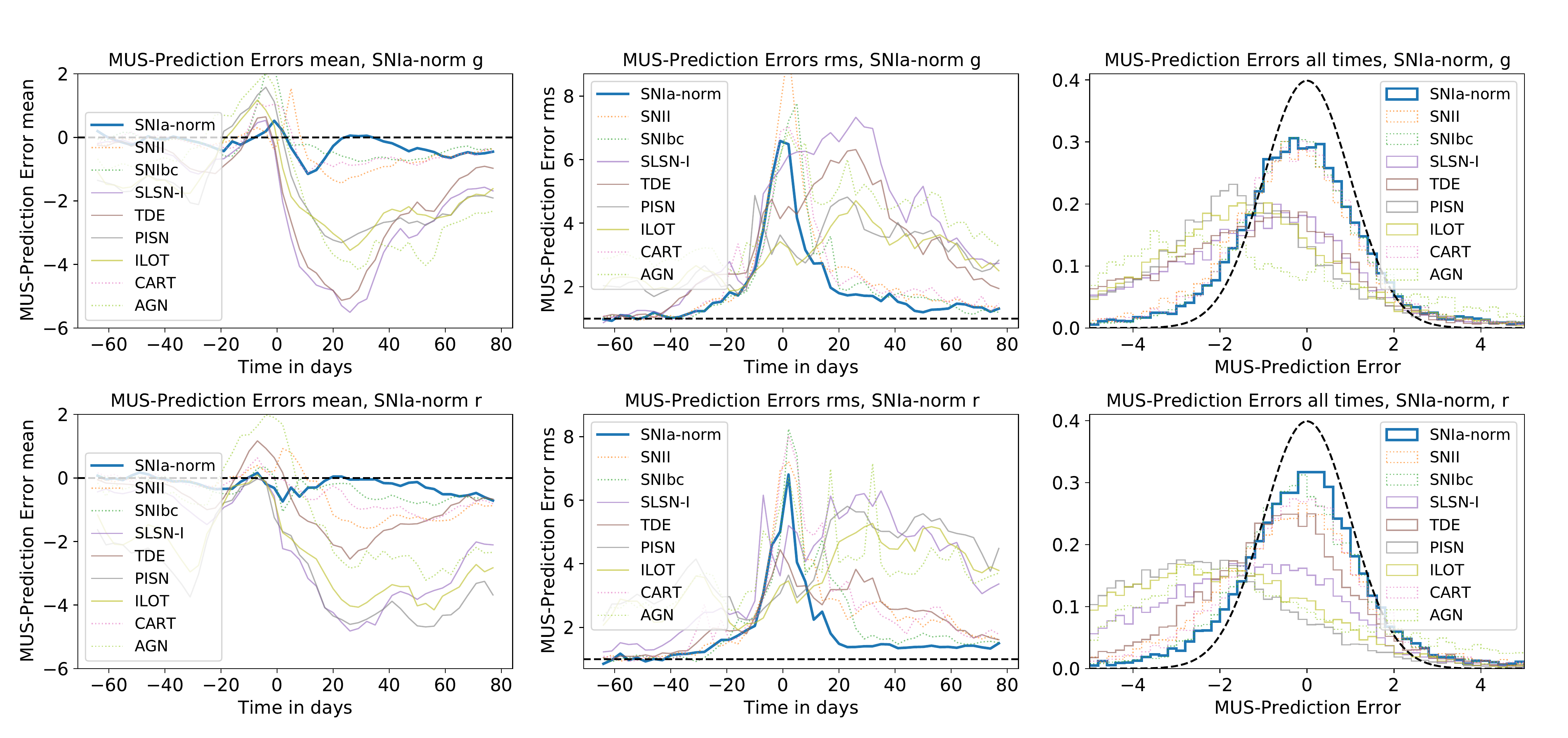}
        \caption{Distribution of the Measurement-Uncertainty-Scaled Prediction Errors (MUSPE) for the Bazin SNIa model at different times. We apply the Bazin SNIa model to each other class' datasets and compute Equation \ref{eq:physical_flux_error} at each time-step. We have not plotted the kilonova or uLens-BSR classes here because they show significant deviations, indicating that the SNIa model is very bad at predicting these classes, and hence easily identifies them as anomalies. We show the mean and root mean square (rms) for each prediction error distribution at each time-step in the first three panels, with the $g$ band shown in the top row of panels, and the $r$ band shown in the bottom row of panels. In the last column, we plot the distribution of scaled errors across all times. We plot a unit Gaussian as a black dashed line to help guide the eye.}
        \label{fig:Bazin_flux_errors}
        \end{figure*}

\begin{figure*}[ht]
        \centering
        \textbf{DNN}\par\medskip
        \vspace{-0.7em}
        \includegraphics[width=1.0\linewidth]{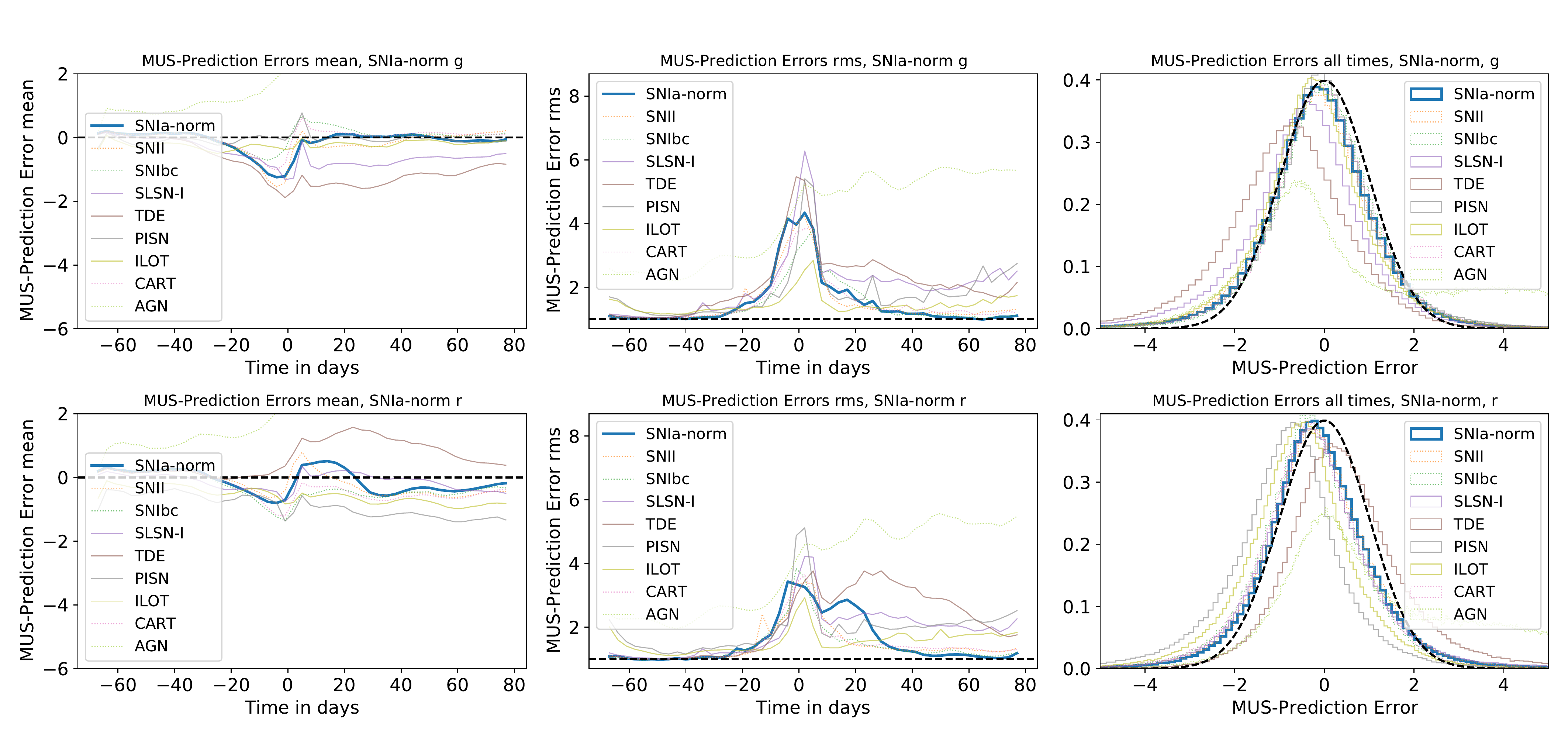}
        \caption{Distribution of the Measurement-Uncertainty-Scaled Prediction Errors (MUSPE) for the DNN SNIa model at different times. We apply the DNN SNIa model to each other class' datasets and compute Equation \ref{eq:physical_flux_error} at each time-step. We have not plotted the kilonova or uLens-BSR classes here because they show significant deviations, indicating that the SNIa model is very bad at predicting these classes, and hence easily identifies them as anomalies. We show the mean and root mean square (rms) for each prediction error distribution at each time-step in the first three panels, with the $g$ band shown in the top row of panels, and the $r$ band shown in the bottom row of panels. In the last column, we plot the distribution of scaled errors across all times. We plot a unit Gaussian as a black dashed line to help guide the eye.}
        \label{fig:DNN_flux_errors}
        \end{figure*}

To analyse why the SNIa Bazin model is better than than the DNN at identifying anomalies, we plot the prediction errors for the SNIa models applied to each of the other transient classes for Bazin and DNN in Figures \ref{fig:Bazin_flux_errors} and \ref{fig:DNN_flux_errors}, respectively. We expect that the SNIa models should predict the SNIa light curves best, and indeed, we see that these blue lines for the SNIa have nearly the best prediction error distributions. In Figure \ref{fig:Bazin_flux_errors}, the prediction errors are significantly worse for the more anomalous classes (SLSNe, TDEs, PISNe, ILOTs) with deviations ranging up to 5 sigma. However, the prediction errors for these classes in the DNN are much smaller, not much more than 1 sigma deviations. This indicates that the Bazin model is much worse at predicting the observations from these anomalous classes than the DNN, and hence is better at identifying them as anomalies despite being better able to predict observations from common transients. We conclude that the flexibility of the DNN makes it excellent at predicting future fluxes for any transient despite only being trained on SNIa. This is great if the task was prediction, but it is a poor choice for anomaly detection in this framework.
        
\end{document}